\documentclass[letterpaper, 10 pt, conference]{ieeeconf}  
\IEEEoverridecommandlockouts                              
\usepackage{packages}
\usepackage{placeins}
\usepackage{xcolor}
\usepackage{pifont}

\definecolor{checkgreen}{RGB}{35,140,70}
\definecolor{crossred}{RGB}{190,55,55}

\newcommand{\greencheck}{\textcolor{checkgreen}{\ding{51}}}
\newcommand{\redcross}{\textcolor{crossred}{\ding{55}}}

\title{\LARGE\bf 

PhoneBot: A Low-Cost Open Humanoid Robot Platform Reusing Smartphones

}

\author{Ruochen Hou$^{1}$, Quanyou Wang$^{1, \dagger}$, Daniel Koh$^{1, \dagger}$, and Dennis W.~Hong$^{1}$\\
\url{https://phonebot.dev}
\thanks{$^{\dagger}$ Co-second authors.}
\thanks{
$^{1}$Robotics and Mechanisms Laboratory (RoMeLa), Department of Mechanical and Aerospace Engineering, University of California, Los Angeles, CA 90095, USA.
{\tt\scriptsize \{houruochen, qywang0628, danielkoh1, dennishong\}@ucla.edu}
}
}

\begin{document}
\maketitle
\thispagestyle{empty}
\pagestyle{empty}

\begin{abstract}
The adoption of humanoid robots in education and research remains limited by high hardware costs, complex sensing systems, and substantial computational requirements. This paper presents PhoneBot, a low-cost, open-source humanoid robot platform that repurposes commodity smartphones as its primary sensing and computing unit. By using a smartphone's integrated inertial measurement unit (IMU), camera, wireless connectivity, and onboard processing capabilities, PhoneBot reduces hardware costs and simplifies the system architecture. The robot combines a modular lower-body structure driven by 13 low-cost actuators with a torso-mounted smartphone that supports perception, control computation, and user interaction. We describe the mechanical design, software architecture, and real-time communication framework that support stable locomotion and capabilities including vision-based human following, conversational interaction, filming, and mobile telepresence. Experimental evaluations demonstrate reliable walking, perception-driven interaction, and straightforward deployment using off-the-shelf consumer smartphones. With fully open-source hardware and software designs, PhoneBot provides an affordable, reproducible platform for education, research, and rapid prototyping. More details are available at \url{https://phonebot.dev}.

\end{abstract}


\section{Introduction}
\label{sec:intro}




Compact humanoid robots provide accessible platforms for studying locomotion, physical interaction, and embodied intelligence. Their reduced size supports experimentation in laboratories and classrooms with limited space and infrastructure. Recent platforms have improved accessibility through open-source hardware, accessible manufacturing methods, and commercially available components \cite{toddlerbot,berkeley_humanoid_lite}. Nevertheless, hardware costs and system integration effort remain barriers to broader adoption in education and research. Beyond mechanical structure assembly, users must connect and configure components for motion control, perception, communication, and interaction. Further reducing these barriers requires simplifying the integration of sensing, computing, and communication capabilities.

As the widely available consumer devices, smartphones offer a practical basis for this approach, where they combine a camera, an IMU, wireless connectivity, onboard processing, and a user interface. Reusing these capabilities can reduce the need for separate sensing and computing hardware while providing an interface for robot operation and interaction. Applying this approach to a humanoid robot requires a system architecture that connects smartphone-based sensing and computation to actuator control and supports both locomotion and interactive behaviors.

\begin{figure} [t]
    \centering
    \includegraphics[width=\linewidth]{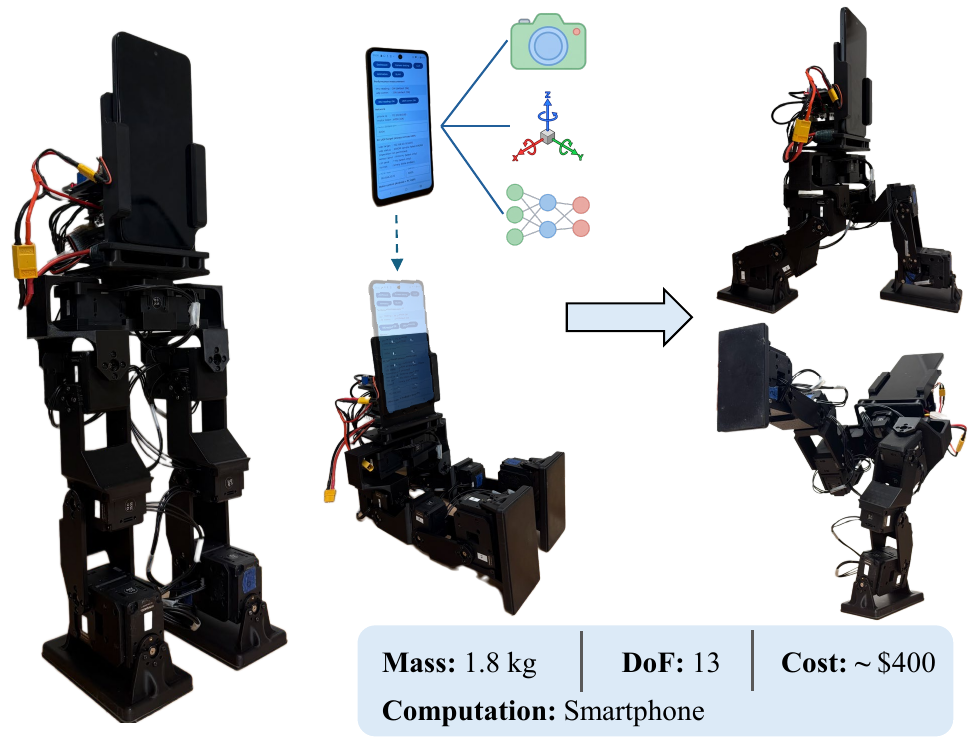}
    \caption{PhoneBot is an open-source platform for bipedal robotics research that uses a commodity smartphone as its onboard computer to reduce hardware cost and simplify system architecture.}
    \label{fig:hero}
\end{figure}

We present PhoneBot (Fig.~\ref{fig:hero}), a low-cost, open-source compact humanoid platform that combines a modular lower body with 13 actuators and a torso-mounted smartphone. By using the smartphone’s integrated functions, PhoneBot consolidates sensing, computation, communication, and user interaction in a single device, reducing the need for separate hardware components. We describe the mechanical design, software architecture, and communication framework that connect smartphone-based perception and control computation to the robot’s actuator control system. The platform supports walking, vision-based human following, fall recovery, and mobile telepresence.

\textbf{Contributions.} The contributions of this work are:
(i) a low-cost, modular, open-source humanoid that reuses smartphone sensing and computation to simplify hardware integration; (ii) a torque-aware locomotion pipeline in which sagittal mirroring produces left--right balanced walking and joint load, and also improves learning speed and consistency under actuator limits; and (iii) an integrated Android application for policy inference, vision, SLAM and human--robot interaction, validated on four old or low-cost smartphones.

\section{Related Works}
\label{related_works}

\begin{table*}[htbp!]
    \centering
    \caption{Comparison with Small-Scale Humanoid Research Platforms.}
    \label{tab:comparison}
    \renewcommand{\arraystretch}{1.1}
    \begin{tabular}{lcccccc}
        \hline
         & \textbf{ROBOTIS OP3}\cite{robotis_op3} & \textbf{ToddlerBot}\cite{toddlerbot} & \textbf{BRUCE}\cite{liu2022bruce} & \textbf{Berkeley Lite}\cite{berkeley_humanoid_lite} & \textbf{HighTorque MiniPi}\cite{mini_pi} & \textbf{Ours} \\
        \hline
        Height (m) & 0.51 & 0.56 & 0.70 & 0.80 & 0.52 & 0.48 \\
        Weight & 3.5 & 3.4 & 4.8 & 16.0 & 7.0 & 1.8 \\
        Active DoFs & 20 & 30 & 16 & 22 & 12 & 13 \\
        Locomotion & \greencheck & \greencheck & \greencheck & \greencheck & \greencheck & \greencheck \\
        Perception & \greencheck & \greencheck & \redcross & \redcross & \redcross & \greencheck \\
        Open Source & Design \& Code & Design \& Code & Code Only & Design \& Code & Code Only & Design \& Code \\
        Price (\$) & 13.8K & 6K & 9.0K & 5.0K & 3.5K & \textbf{$\sim400$}$^{\dagger}$ \\
        \hline
    \end{tabular}
    
    \vspace{2pt}
    {\footnotesize $^{\dagger}$Price excludes the smartphone.}
\end{table*}

Compact humanoids support research in locomotion, control, and human-robot interaction. ROBOTIS OP3 is a miniature humanoid platform with integrated actuators, sensors, and an onboard computer~\cite{robotis_op3}. BRUCE emphasizes dynamic motion through proprioceptive actuation and mechanisms that reduce leg inertia~\cite{liu2022bruce}, while Mini Pi provides a compact bipedal platform designed for robust locomotion~\cite{mini_pi}. Other platforms emphasize hardware accessibility and reproducibility. ToddlerBot combines a 3D-printed structure and commercially available actuators with tools for learning-based locomotion and manipulation~\cite{toddlerbot}. Berkeley Humanoid Lite uses modular 3D-printed gearboxes and widely available components to reduce manufacturing costs~\cite{berkeley_humanoid_lite}. These platforms demonstrate complementary approaches to balancing capability, cost, and reproducibility. 

Smartphones integrate computation, sensing, wireless communication, and user interfaces in a single device. OpenBot uses these capabilities by pairing an Android smartphone with a low-cost wheeled body for onboard perception and autonomous navigation~\cite{openbot}. PhoneBot shares the objective of reusing smartphone capabilities to reduce dedicated hardware requirements, while integrating them into a compact humanoid platform for locomotion and interaction.

Humanoid locomotion has been studied through both model-based and learning-based approaches. Model-based methods use robot dynamics for motion planning, tracking, and balance control~\cite{liu2022bruce,liu2025kidcosmo,scianca2020mpc}. Reinforcement learning has enabled diverse bipedal locomotion behaviors, including walking, running, and jumping~\cite{li2025reinforcement}. Successful simulation-to-real transfer depends on both policy design and the correspondence between simulated and physical robots~\cite{gu2024humanoid}. Existing approaches include transformer policies trained in randomized simulation~\cite{radosavovic2024real}, hardware designs that reduce simulation complexity~\cite{liao2025berkeley}, and explicit modeling of parallel actuation mechanisms during policy training~\cite{tanaka2025mechanical}. 

GPU-accelerated simulation improves training efficiency by enabling large-scale parallel rollout collection~\cite{rudin2022learning,mittal2023orbit,zakka2025mujoco}. Domain randomization and asymmetric actor--critic training have been used to support sim-to-real transfer~\cite{nahrendra2023dreamwaq}, while symmetry-based data augmentation can improve learning efficiency and gait symmetry~\cite{mittal2024symmetry, su2024leveraging}. 
 
\section{Hardware Design}
\label{sec:hardware}
\begin{figure} [t]
    \centering
    \includegraphics[width=\linewidth]{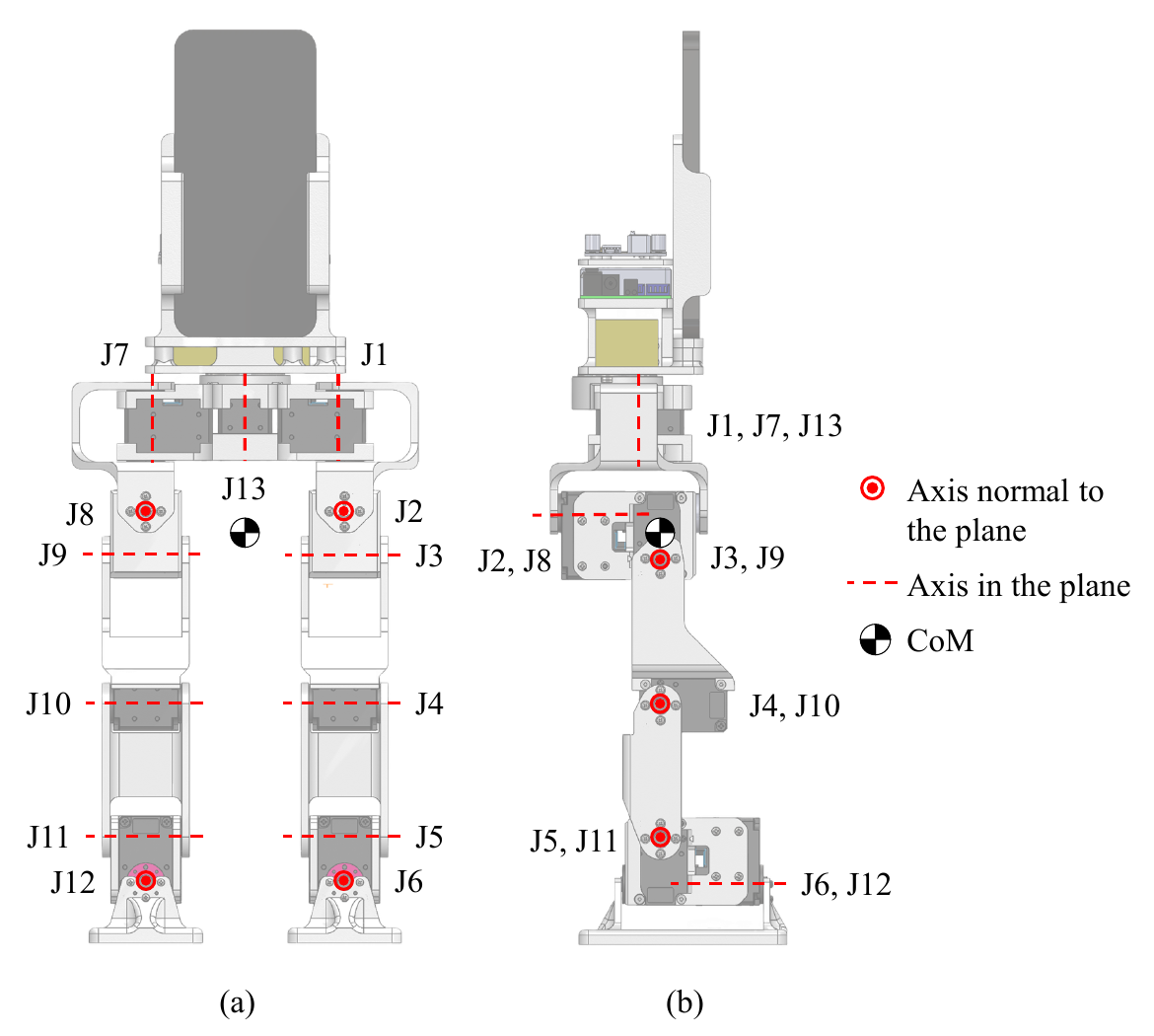}
    \caption{Joint configuration and CoM of PhoneBot. (a) Front view. (b) Left view.}
    \label{fig:configuration}
\end{figure}

\begin{table}[t]
    \centering
    \caption{Joint specifications and Dynamixel motor assignments.}
    \label{tab:joint_specs}
    \renewcommand{\arraystretch}{1.1}
    \begin{tabular}{lccc}
        \hline
        Name & Joint & Actuator & Range ($^\circ$) \\
        \hline
        \noalign{\vskip 1pt}
        Hip yaw     & J1, J7  & XL430-W250 & $[-90,\,90]$ \\
        Hip roll    & J2, J8  & XL430-W250 & $[-90,\,90]$ \\
        \multirow{2}{*}{Hip pitch}
                    & J3      & XL430-W250 & $[-90,\,123]$ \\
                    & J9      & XL430-W250 & $[-123,\,90]$ \\
        Knee pitch  & J4, J10 & XL430-W250 & $[-90,\,90]$ \\
        Ankle pitch & J5, J11 & XL430-W250 & $[-90,\,90]$ \\
        Ankle roll  & J6, J12 & XL430-W250 & $[-90,\,90]$ \\
        Torso yaw   & J13     & XL430-W250 & $[-90,\,90]$ \\
        \hline
    \end{tabular}
\end{table}

\begin{table}[t]
    \centering
    \caption{Manufacturer specifications of the DYNAMIXEL XL430-W250 actuator.}
    \label{tab:motor_specs}
    \renewcommand{\arraystretch}{1.1}
    \begin{tabular}{cccc}
        \toprule
        \shortstack{\textbf{Gear Ratio} \\ \strut} & \shortstack{\textbf{Stall Torque}\\(N\,m)} & \shortstack{\textbf{No-Load Speed}\\(RPM)} & \shortstack{\textbf{Stall Current}\\(A)} \\
        \midrule
        $258.5{:}1$ & $1.4$ & $57$ & $1.3$ \\
        \toprule
        \shortstack{\textbf{Voltage}\\(V)} & \shortstack{\textbf{Resolution}\\(Pulse/Rev)} & \shortstack{\textbf{Mass}\\(g)} & \shortstack{\textbf{Dimensions}\\(mm)} \\
        \midrule
        $11.1$ & $4096$ & $57.2$ &  $46.5\times34.0\times28.5$ \\
        \bottomrule
    \end{tabular}
\end{table}

\begin{figure} [t]
    \centering
    \includegraphics[width=\linewidth]{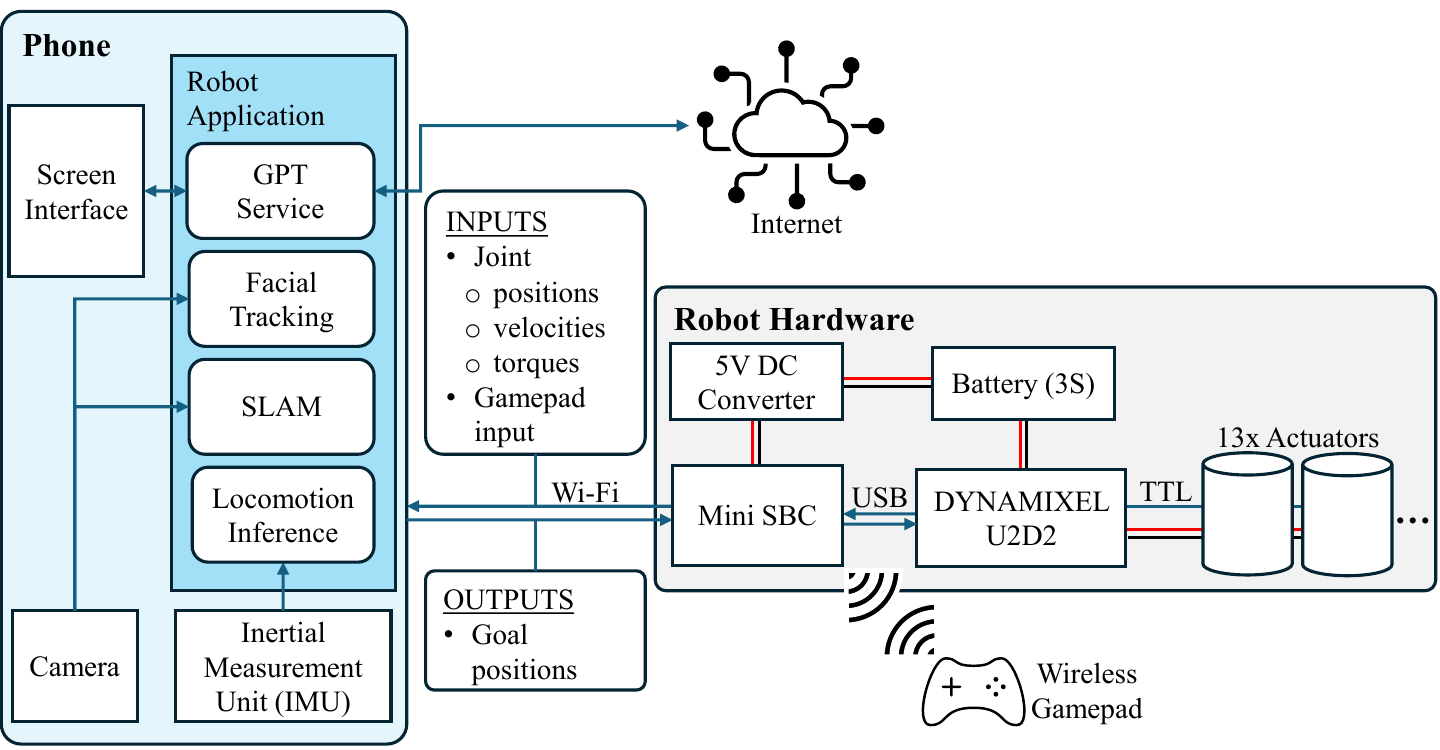}
    \caption{System overview of PhoneBot}
    \label{fig:system_diagram}
\end{figure}


The hardware design of PhoneBot emphasizes low cost, modularity, and compact integration while maintaining sufficient mobility for dynamic bipedal locomotion and vision-based interaction. A commodity smartphone serves as both the primary sensing platform and the high-level computing unit, integrating the camera, inertial sensing, wireless communication, and application-level computation within a single readily available device. This arrangement reduces the need for dedicated sensor hardware and concentrates most of the mechanical and actuation requirements in the lower-body locomotion system. 

Table~\ref{tab:comparison} summarizes PhoneBot's physical specifications and shows the comparison with other popular small-scale humanoid platforms. The assembled platform measures $483 \times 183 \times 125~\mathrm{mm}$ ($\mathrm{H}\times\mathrm{W}\times\mathrm{D}$) and has a total mass of $1.8~\mathrm{kg}$, including the smartphone, actuators, structural components, electronics, and battery. The relatively narrow body profile and symmetric lower-body construction are intended to provide a compact platform suitable for both locomotion and mobile perception experiments.

\subsection{Kinematic Structure and Joint Configuration}
\label{subsec:kinematics}

The kinematic topology and center-of-mass (CoM) location are illustrated in Fig.~\ref{fig:configuration}. PhoneBot has a total of 13 degrees of freedom (DoF), comprising 6 actuated joints per leg arranged in a conventional serial biped configuration: hip yaw, hip roll, hip pitch, knee pitch, ankle pitch, and ankle roll. The right and left legs are actuated by joints J1--J6 and J7--J12, respectively, as summarized in Table~\ref{tab:joint_specs}. This symmetric arrangement provides 3 rotational DoF at each hip, 1 DoF at each knee, and 2 DoF at each ankle, enabling control of foot placement and orientation in three-dimensional space. An additional torso yaw joint, J13, connects the lower-body assembly to the smartphone mounting structure. This joint allows the upper sensing platform to rotate independently of the stance direction, enabling active visual tracking, camera reorientation, and perception tasks without requiring the robot to rotate its entire body.

As shown in Table~\ref{tab:joint_specs}, most joints are mechanically constrained to an operating range of $[-90^{\circ},90^{\circ}]$. The hip pitch joints, J3 and J9, have an asymmetric extended range. The larger hip-pitch excursion increases the available leg workspace in the sagittal plane and permits greater thigh flexion during configurations such as deep crouching, large forward leg swings, and recovery motions. All 13 joints are actuated using DYNAMIXEL XL430-W250-T servo motors, with key specifications summarized in Table~\ref{tab:motor_specs}. At $\$23.90$ per unit, this actuator minimizes platform cost despite being the system's primary expense.

\subsection{Electronic and System Integration}
\label{subsec:system_integration}

Fig.~\ref{fig:system_diagram} illustrates the system structure of PhoneBot. PhoneBot uses the smartphone for perception, locomotion policy inference, and access to cloud-connected services, while the onboard mini single-board computer (SBC) handles communication with the actuators and execution of joint-position commands. Utilizing the smartphone's integrated camera, IMU, and wireless interfaces eliminates the need to source standalone modules that require precision mounting and calibration.

The smartphone has no physical electrical connection with the robot hardware. Instead, high-level control commands, joint telemetry, and gamepad inputs are exchanged bidirectionally over Wi-Fi between the smartphone and the SBC. This is an intentional design choice to overcome hardware limitations in wired connections, and allow for compatibility with different smartphone models capable of running the software application.

\section{Locomotion}
\label{sec:locomotion}

\subsection{Torque-Aware Actuation Model}
\label{sec:actuation}

The actuators on PhoneBot are torque-limited (stall torque $=1.4~\mathrm{N{\cdot}m}$), therefore a position actuator that ignores the torque--speed envelope overestimates available force, especially near no-load speed. Using ToddlerBot's torque-limited Dynamixel model as a reference~\cite{toddlerbot}, we convert each policy position command to a saturated torque in MuJoCo/MJX.

Let $q$, $\dot{q}$, and $\ddot{q}$ denote joint position, velocity, and acceleration. The clipped position target includes an episode-constant offset $\delta$:
\begin{equation}
\hat{q}
=
\mathrm{clip}_{[\underline{q},\overline{q}]}
\!\left(q_{\mathrm{home}} + a + \delta\right),
\qquad
\delta \sim \mathcal{U}[-0.05,\,0.05]~\mathrm{rad}.
\label{eq:motor_target}
\end{equation}
The unconstrained PD torque is
\begin{equation}
\tau_{m}
=
k_{p}^{\mathrm{eff}}\,(\hat{q}-q)
-
\bigl(k_{d}^{\min}+k_{d}\bigr)\,\dot{q}.
\label{eq:pd_torque}
\end{equation}
Firmware gains are converted to simulator units by $k_{p}=k_{p}^{\mathrm{real}}/r_{p}$ and $k_{d}=k_{d}^{\mathrm{real}}/r_{d}$. When the motor is back-driven ($\ddot{q}\,(\hat{q}-q)<0$), gearbox inefficiency is modeled by increasing stiffness through a passive--active ratio $\rho$:
\begin{equation}
k_{p}^{\mathrm{eff}}
=
\begin{cases}
\rho\,k_{p}, & \ddot{q}\,(\hat{q}-q)<0,\\
k_{p}, & \text{otherwise}.
\end{cases}
\label{eq:kp_eff}
\end{equation}

The acceleration-side torque limit is a piecewise-linear function of speed. It stays at $\tau_{\max}$ up to $\dot{q}_{\tau_{\max}}$, then tapers to a residual $\tau_{\dot{q}_{\max}}$ at $\dot{q}_{\max}$:
\begin{equation}
\tau_{\mathrm{acc}}(\dot{q})
=
\begin{cases}
\tau_{\max},
& |\dot{q}| \le \dot{q}_{\tau_{\max}},\\[2pt]
\tau_{\max}
+
\dfrac{\tau_{\dot{q}_{\max}}-\tau_{\max}}{\dot{q}_{\max}-\dot{q}_{\tau_{\max}}}
\bigl(|\dot{q}|-\dot{q}_{\tau_{\max}}\bigr),
& |\dot{q}| > \dot{q}_{\tau_{\max}}.
\end{cases}
\label{eq:tau_acc}
\end{equation}
Braking torque is a larger constant $\tau_{\mathrm{brake}}$. The applied command is the asymmetric clamp
\begin{equation}
\tau_{\mathrm{clip}}
=
\begin{cases}
\mathrm{clip}\bigl(\tau_{m},\,-\tau_{\mathrm{brake}},\,\tau_{\mathrm{acc}}\bigr),
& \dot{q} \ge 0,\\
\mathrm{clip}\bigl(\tau_{m},\,-\tau_{\mathrm{acc}},\,\tau_{\mathrm{brake}}\bigr),
& \dot{q} < 0.
\end{cases}
\label{eq:tau_clip}
\end{equation}
An accelerating command above $\dot{q}_{\max}$ triggers full braking, $\tau=-\mathrm{sign}(\dot{q})\tau_{\mathrm{brake}}$. MuJoCo adds passive torque $\tau_p=-d\dot{q}-\tau_f\mathrm{sign}(\dot{q})$ and armature $I$ (Table~\ref{tab:xl430_actuation}).

Encoder backlash is applied only to the observed joint position used by the policy,
\begin{equation}
\tilde{q}
=
q
+
\tfrac{1}{2}\,b\,\tanh(\tau/\alpha),
\label{eq:backlash}
\end{equation}
where the backlash width $b$ is sampled once per episode from $\mathcal{U}[0.02,\,0.1]~\mathrm{rad}$ and $\alpha=0.1~\mathrm{N{\cdot}m}$ sets the load at which the offset saturates. Nominal XL430 parameters are listed in Table~\ref{tab:xl430_actuation}. During training the envelope gains are further multiplied by per-episode factors ($k_{p},k_{d}$ in $[0.7,1.3]$, $\tau_{\max},\tau_{\mathrm{brake}},\tau_{\dot{q}_{\max}}$ in $[0.75,1.1]$, speed breakpoints and $k_{d}^{\min}$ in $[0.9,1.1]$, and $\rho$ in $[0.5,1.5]$).

\begin{table}[t]
\centering
\caption{Dynamixel XL430 actuation parameters used in the torque-aware simulator. All 13 joints share this model.}
\label{tab:xl430_actuation}
\renewcommand{\arraystretch}{1.15}
\begin{tabular}{lcc}
\toprule
Parameter & Symbol & Value \\
\midrule
Firmware / sim $k_{p}$ scale & $r_{p}$ & $150$ \\
Firmware / sim $k_{d}$ scale & $r_{d}$ & $16$ \\
Firmware position gain & $k_{p}^{\mathrm{real}}$ & $600$ \\
Firmware velocity gain & $k_{d}^{\mathrm{real}}$ & $0$ \\
Simulator stiffness & $k_{p}$ & $4.0~\mathrm{N{\cdot}m/rad}$ \\
Simulator damping & $k_{d}$ & $0$ \\
Powered-on extra damping & $k_{d}^{\min}$ & $0.162~\mathrm{N{\cdot}m{\cdot}s/rad}$ \\
Passive--active ratio & $\rho$ & $3.0$ \\
\midrule
Peak accel.\ torque & $\tau_{\max}$ & $0.93~\mathrm{N{\cdot}m}$ \\
Start of torque taper & $\dot{q}_{\tau_{\max}}$ & $2.0~\mathrm{rad/s}$ \\
No-load speed & $\dot{q}_{\max}$ & $5.97~\mathrm{rad/s}$ \\
Residual accel.\ torque & $\tau_{\dot{q}_{\max}}$ & $0.08~\mathrm{N{\cdot}m}$ \\
Brake torque & $\tau_{\mathrm{brake}}$ & $1.4~\mathrm{N{\cdot}m}$ \\
\midrule
Joint damping & $d$ & $0.0013~\mathrm{N{\cdot}m{\cdot}s/rad}$ \\
Rotor armature & $I$ & $0.0083~\mathrm{kg{\cdot}m^{2}}$ \\
Friction loss & $\tau_{f}$ & $0.069~\mathrm{N{\cdot}m}$ \\
Command offset range & $\delta$ & $[-0.05,\,0.05]~\mathrm{rad}$ \\
Backlash width range & $b$ & $[0.02,\,0.1]~\mathrm{rad}$ \\
Backlash activation & $\alpha$ & $0.1~\mathrm{N{\cdot}m}$ \\
\bottomrule
\end{tabular}
\end{table}

\subsection{Observation, Action, and Reward}
\label{sec:walk_policy}

We train an omnidirectional walk policy with asymmetric actor--critic PPO. The actor receives only signals that exist on the phone (noisy IMU and joint feedback). The critic receives a larger privileged state that exists only in simulation, which improves value estimation without changing the observation used at deployment.

\paragraph{Observation space.}
The actor observation $o_t$ concatenates noisy gyroscope $\omega_t$, projected gravity $g_t$, the velocity command $c_t=(v_x,v_y,\omega_z)$, joint positions relative to the offset home pose, joint velocities, the previous action, and a two-foot gait phase $(\cos\phi_L,\cos\phi_R,\sin\phi_L,\sin\phi_R)$. Joint positions include the load-dependent backlash of~\eq{eq:backlash}. When $\|c_t\|$ is below a deadzone of $0.05$, the phase provided to the networks is replaced by $(\pi,\pi)$ so the same policy can stand still. The critic input is this proprioceptive vector plus a noisy base linear velocity, a noiseless copy of IMU and joint state, and quantities available only in simulation: base height, actuator force, binary foot contacts, foot linear velocities, and swing air time. Table~\ref{tab:obs_space} lists this partition. Actor dimension is $52$; critic dimension is $120$.

\begin{table}[t]
\centering
\caption{Observation space of the torque-aware walk policy. Actor inputs are the onboard signals used at deployment. Critic inputs add privileged simulation state.}
\label{tab:obs_space}
\renewcommand{\arraystretch}{1.1}
\footnotesize
\begin{tabular}{lccc}
\toprule
Component & Dim. & Actor & Critic \\
\midrule
Angular velocity (noisy) & $3$ & $\checkmark$ & $\checkmark$ \\
Gravity vector (noisy) & $3$ & $\checkmark$ & $\checkmark$ \\
Velocity command & $3$ & $\checkmark$ & $\checkmark$ \\
Joint position (noisy, rel.\ home) & $13$ & $\checkmark$ & $\checkmark$ \\
Joint velocity (noisy) & $13$ & $\checkmark$ & $\checkmark$ \\
Previous action & $13$ & $\checkmark$ & $\checkmark$ \\
Gait phase $(\cos,\sin)$ & $4$ & $\checkmark$ & $\checkmark$ \\
\midrule
Linear velocity (noisy) & $3$ & & $\checkmark$ \\
Angular velocity (clean) & $3$ & & $\checkmark$ \\
Accelerometer & $3$ & & $\checkmark$ \\
Gravity vector (clean) & $3$ & & $\checkmark$ \\
Linear velocity (clean) & $3$ & & $\checkmark$ \\
Global angular velocity & $3$ & & $\checkmark$ \\
Joint position (clean) & $13$ & & $\checkmark$ \\
Joint velocity (clean) & $13$ & & $\checkmark$ \\
Base height & $1$ & & $\checkmark$ \\
Actuator force & $13$ & & $\checkmark$ \\
Foot contact & $2$ & & $\checkmark$ \\
Foot linear velocity & $6$ & & $\checkmark$ \\
Foot air time & $2$ & & $\checkmark$ \\
\midrule
Total & & $52$ & $120$ \\
\bottomrule
\end{tabular}
\end{table}

\paragraph{Action space.}
The policy outputs a $13$-dimensional joint-position residual $a_t$ at $50~\mathrm{Hz}$. Desired positions follow~\eq{eq:motor_target} and are converted to torque by the XL430 model of Sec.~\ref{sec:actuation}, not by an unsaturated position actuator. Commands are sampled in $v_x,v_y\in[-0.5,0.5]~\mathrm{m/s}$ and $\omega_z\in[-1.0,1.0]~\mathrm{rad/s}$, with probability $0.1$ of a zero (stand) command.

\paragraph{Reward function.}
The scalar per-step reward is $r=\Delta t\sum_i w_i r_i$ with control timestep $\Delta t=0.02~\mathrm{s}$ ($50~\mathrm{Hz}$). Tracking terms reward matching IMU-local $v_{xy}$ and $\omega_z$ to $c_t$. Gait terms encourage alternating swing, a $3~\mathrm{cm}$ foot-height schedule, feet parallel to the ground, and zero foot slip. Regularization penalizes XL430 torque near the saturation envelope: $\ell_1$ and $\ell_2$ torque costs plus a utilization penalty that is nonzero only when $|\tau|$ exceeds $80\%$ of the directional limit from~\eq{eq:tau_clip}. Pose costs use a weighted squared error with $w=0.5$ on hip pitch, knee, and ankle and $w=1$ elsewhere, so those joints may deviate from the home pose during stepping. Table~\ref{tab:walk_rewards} lists all terms with nonzero weights.

\begin{table}[t]
\centering
\caption{Reward terms and weights of the torque-aware walk policy. Velocities are IMU-local. $\sigma=0.1$, $h^*=0.03~\mathrm{m}$, $u=|\tau|/\tau_{\mathrm{lim}}$, and $\mathbb{I}_{\mathrm{move}}=\mathbf{1}_{\|c\|>0.05}$.}
\label{tab:walk_rewards}
\renewcommand{\arraystretch}{1.12}
\setlength{\tabcolsep}{3pt}
\footnotesize
\begin{tabular}{@{}llc@{}}
\toprule
Term & Equation & Weight \\
\midrule
Linear velocity tracking & $\exp\bigl(-\|v_{xy}-c_{xy}\|_2^2/\sigma\bigr)$ & $4.0$ \\
Yaw velocity tracking & $\exp\bigl(-(\omega_z-c_z)^2/\sigma\bigr)$ & $4.0$ \\
Alive & $1$ & $0.5$ \\
\midrule
Vertical velocity & $v_z^2$ & $-0.2$ \\
Roll/pitch rate & $\|\omega_{xy}\|_2^2$ & $-0.1$ \\
Orientation & $\|g_{xy}\|_2^2$ & $-10.0$ \\
Torque $\ell_1$ & $\sum_i|\tau_i|$ & $-0.1$ \\
Torque $\ell_2$ & $\mathrm{mean}(\tau^2)$ & $-1.0$ \\
Torque utilization & $\mathrm{mean}\bigl(\mathrm{clip}((u-0.8)/0.2,0,1)^2\bigr)$ & $-2.0$ \\
Action rate & $\|a_t-a_{t-1}\|_2^2$ & $-0.2$ \\
Joint acceleration & $\|\ddot{q}\|_2^2$ & $-2{\times}10^{-6}$ \\
Joint velocity & $\|\dot{q}\|_2^2$ & $-0.002$ \\
Joint limit & $\sum_i[\underline{q}_i-q_i]_+ + [q_i-\overline{q}_i]_+$ & $-1.0$ \\
Pose to home & $\sum_i w_i(q_i-q_i^0)^2$ & $-1.5$ \\
Hip roll deviation & $\sum |q_{\mathrm{hip}}-q_{\mathrm{hip}}^0|\,\mathbf{1}_{|c_y|<0.1}$ & $-1.0$ \\
Knee deviation & $\sum |q_{\mathrm{knee}}-q_{\mathrm{knee}}^0|$ & $-0.005$ \\
Stand still & $\sum |q-q^0|\,\mathbf{1}_{\|c\|\le 0.05}$ & $-0.1$ \\
\midrule
Feet air time & $\sum (t_{\mathrm{air}}-0.2)_+\,\mathbb{I}_{\mathrm{first}}\,\mathbb{I}_{\mathrm{move}}$ & $10.0$ \\
Feet phase & $\exp(-\|z-z^\star(\phi)\|_2^2/\sigma_\phi)\,\mathbb{I}_{\mathrm{move}}$ & $1.2$ \\
Feet corner phase & $\exp(-\|z_{\mathrm{c}}-z^\star(\phi)\|_2^2/\sigma_{\mathrm{c}})\,\mathbb{I}_{\mathrm{move}}$ & $1.5$ \\
Feet clearance & $\sum |z-h^*|\sqrt{\|v_{\mathrm{feet},xy}\|}$ & $-5.0$ \\
Feet height & $\sum ((z_{\mathrm{peak}}/h^*-1)^2)\,\mathbb{I}_{\mathrm{first}}$ & $-5.0$ \\
Feet corner clearance & $\sum [h^*-z_{\mathrm{c,min}}]_+\,\sqrt{\|v_{\mathrm{feet},xy}\|}$ & $-10.0$ \\
Feet parallel & $\sum \|e_{z,\mathrm{foot},xy}\|_2^2$ & $-5.0$ \\
Feet slip & $\sum \|v_{\mathrm{feet},xy}\|_2^2\,\mathbf{1}_{\mathrm{contact}}$ & $-10.0$ \\
Feet distance & $[0.05-d_{\mathrm{feet}}]_+$ & $-1.0$ \\
Foot collision & $\mathbf{1}_{\mathrm{L\text{-}R\ contact}}$ & $-1.0$ \\
\bottomrule
\end{tabular}
\end{table}

\subsection{Sagittal Mirroring}
\label{sec:mirroring}

On a torque-limited platform the walk policy can converge to an asymmetric local optimum: one foot remains forward, the other remains behind, and most of the load and XL430 torque are applied through a single support leg. We observed this asymmetric gait, with the left foot leading and the right foot supporting, on checkpoints trained without mirroring. PhoneBot is left--right symmetric, so that gait is an avoidable bias rather than a mechanical necessity.

We therefore augment each PPO update with a sagittal-plane mirror of the collected batch. One physics step produces the real transition; the trainer then concatenates a mirrored copy $(\,o'_t,a'_t\,)$ for the actor loss. Advantages and returns are computed only on the real trajectory and repeated for the mirrored policy samples, so the critic does not estimate values for synthetic trajectories. The value loss is computed only on real samples: the privileged critic contains world-frame signals that cannot be mirrored without the floating-base yaw.

Mirroring is defined in the IMU frame ($+x$ forward, $+y$ left, $+z$ up). Polar vectors (gravity, linear velocity, acceleration) are scaled by $\mathrm{diag}(1,-1,1)$; angular rates by $\mathrm{diag}(-1,1,-1)$; and the command $(v_x,v_y,\omega_z)$ by $\mathrm{diag}(1,-1,-1)$. Left and right legs are swapped, the gait-phase pair is swapped, and every hinge angle---including trunk yaw---is negated. The last negation follows the PhoneBot CAD: every joint axis is local $+z$, so after the left/right permute a sagittal reflection reverses the rotation sign on all $13$ actuators. The same permutation and signs are applied to $a_t$ and to the Gaussian policy mean, while the policy scale is only permuted.

The result is that a gait that favors the left leg is trained as if a gait that favors the right leg had also been simulated, without a second simulation step. After this augmentation is enabled, the walk no longer depends on a single support leg. Because each physics step provides two policy samples, tracking rewards also increase in fewer environment steps than training without mirroring. Section~\ref{sec:experiments} reports this efficiency with training curves and measures left--right balance with mean torque difference and mean joint-angle difference between the two legs.

\subsection{Standup and Heading Policies}
\label{sec:standup_heading}

A standup policy is trained to recover from sitting and prone poses to the home standing height $z^\star=0.282~\mathrm{m}$. A heading policy holds that stance and aims the phone camera at a world look direction. Both use the same XL430 torque model, the same $13$-dimensional action, and the same actor/critic observation split as walk, except that the three command numbers are remaining look error in the IMU frame, $(\sin\theta,\cos\theta,\varphi)$, rather than $(v_x,v_y,\omega_z)$. Heading starts a walk gait phase when the look yaw in the pelvis frame exceeds $55^\circ$ and stops it below $30^\circ$, so the robot steps only when trunk yaw cannot finish the turn.

Table~\ref{tab:standup_rewards} lists the standup rewards. Pose-to-home terms are weak so that a recumbent robot is not paid for folding the legs while still on the floor. Height, upward velocity, and upright orientation lead the get-up. A stable-stand bonus is paid only when height, tilt, pose, and base velocity are all near the standing target. Extra action-rate and foot-slip costs apply only near that stand. Trunk-to-floor contact is always penalized. Table~\ref{tab:heading_rewards} lists heading-specific terms. While the heading policy is stepping, it also uses the walk gait terms in Table~\ref{tab:walk_rewards} at the same weights; those gait terms are zero while the robot is planted.

\begin{table}[t]
\centering
\caption{Standup reward terms and weights. The target height is $z^\star=0.282~\mathrm{m}$. The terms $h_c$, $u_c$, and $p_c$ are Gaussian height, uprightness, and pose scores; each $\sigma$ is the width of its corresponding Gaussian.}
\label{tab:standup_rewards}
\renewcommand{\arraystretch}{1.05}
\setlength{\tabcolsep}{2.5pt}
\footnotesize
\begin{tabular}{@{}llc@{}}
\toprule
Term & Equation & Weight \\
\midrule
Height & $\exp(-(z-z^\star)^2/\sigma_z^2)$ & $1.0$ \\
Height (sharp) & $\exp(-(z-z^\star)^2/\sigma_{z,\mathrm{s}}^2)$ & $1.0$ \\
Height $\ell_1$ & $-|z-z^\star|$ & $7.5$ \\
Upward velocity & $\max(v_z,0)\,\mathrm{clip}((z^\star-z)/0.02,0,1)$ & $0.75$ \\
Upright linear & $g_z$ & $1.5$ \\
Upright sharp & $\exp(-(\|g_{xy}\|/\sigma_u)^2)$ gated by height & $1.5$ \\
Standing product & $h_c\,u_c\,p_c$ & $0.375$ \\
Stable stand & $h_c\,u_c\,p_c\exp(-(v_z/\sigma_v)^2-\|v_{xy}\|_2^2/\sigma_v^2)$ & $0.5$ \\
Pose (Gaussian mean) & $\mathrm{mean}_i\exp(-((q_i-q_i^0)/\sigma_q)^2)$ & $0.02$ \\
Pose $\ell_1$ & $-\mathrm{mean}_i|q_i-q_i^0|$ & $0.0125$ \\
Gentle rise & $-|a_z|$ & $0.005$ \\
Action rate & $\|a_t-a_{t-1}\|_2^2$ & $-0.1$ \\
Action rate (stand) & $\|a_t-a_{t-1}\|_2^2\,h_c u_c$ & $-0.4$ \\
Roll/pitch rate & $\|\omega_{xy}\|_2^2$ & $-0.05$ \\
Feet slip (stand) & $\sum \|v_{\mathrm{feet},xy}\|_2^2\,\mathbf{1}_{\mathrm{contact}}\,h_c u_c$ & $-5.0$ \\
Self-collision & $n_{\mathrm{coll}}$ & $-1.0$ \\
Joint limit & $\sum_i[\underline{q}_i-q_i]_+ + [q_i-\overline{q}_i]_+$ & $-1.0$ \\
Trunk--floor & $\mathbf{1}_{\mathrm{trunk\text{-}floor}}$ & $-1.0$ \\
\bottomrule
\end{tabular}
\end{table}

\begin{table}[t]
\centering
\caption{Heading reward terms and weights. The terms $\theta$ and $\varphi$ are remaining look yaw and elevation in the IMU frame, and $\psi$ is look yaw in the pelvis frame. Gait terms while stepping match Table~\ref{tab:walk_rewards}.}
\label{tab:heading_rewards}
\renewcommand{\arraystretch}{1.1}
\setlength{\tabcolsep}{1.5pt}
\footnotesize
\begin{tabular}{@{}llc@{}}
\toprule
Term & Equation & Weight \\
\midrule
Look yaw & $\exp(-(\theta/\sigma_\theta)^2)$ & $1.5$ \\
Look elevation & $\exp(-(\varphi/\sigma_\varphi)^2)$ & $1.0$ \\
Look product & $r_{\mathrm{yaw}}\,r_{\mathrm{elev}}$ & $1.0$ \\
Pelvis look yaw & $\exp(-(\psi/\sigma_\theta)^2)$ & $2.0$ \\
Zero IMU $v_{xy}$ & $\exp(-\|v_{xy}\|_2^2/\sigma)\,(1-\mathbb{I}_{\mathrm{step}})$ & $4.0$ \\
Stepping tax & $\mathbb{I}_{\mathrm{step}}$ & $-0.5$ \\
Height & $\exp(-(z-z^\star)^2/\sigma_z^2)$ & $1.0$ \\
Height $\ell_1$ & $|z-z^\star|$ & $-4.0$ \\
Pose to home & $\sum_i w_i(q_i-q_i^0)^2$ & $-1.5$ \\
Action rate & $\|a_t-a_{t-1}\|_2^2$ & $-0.5$ \\
Roll/pitch rate & $\|\omega_{xy}\|_2^2$ & $-0.25$ \\
Torque $\ell_1$ / $\ell_2$ & $\sum|\tau_i|$ / $\mathrm{mean}(\tau^2)$ & $-0.1$ / $-1.0$ \\
Torque utilization & same as Table~\ref{tab:walk_rewards} & $-2.0$ \\
Joint vel.\ / acc. & $\|\dot{q}\|_2^2$ / $\|\ddot{q}\|_2^2$ & $-0.002$ / $-2{\times}10^{-6}$ \\
Feet slip & $\sum \|v_{\mathrm{feet},xy}\|_2^2\,\mathbf{1}_{\mathrm{contact}}$ & $-10.0$ \\
Self-collision / limits & $n_{\mathrm{coll}}$ / joint-limit cost & $-1.0$ / $-1.0$ \\
\bottomrule
\end{tabular}
\end{table}

\section{Smartphone Application}
\label{sec:application}

PhoneBot uses a commodity smartphone as the torso, the sensor suite, and the onboard computer. OpenBot showed that a standard Android phone already provides cameras, an IMU, CPU and GPU compute, and wireless communication, and that this hardware is sufficient for robotics workloads such as person following and navigation when it is attached to a low-cost wheeled body \cite{openbot}. PhoneBot applies the same idea to a biped. The phone is fixed to the trunk, so its IMU is the body IMU used by the locomotion policies, and its camera is the robot camera.

A smartphone also provides a maintained software platform (Android or iOS), yearly hardware updates, and a second-hand market. The electromechanical legs do not need a separate vision computer or a custom IMU board. The current PhoneBot application is an Android app that uses these built-in devices directly.

\paragraph{Inertial sensing and locomotion.}
The app reads the phone IMU at the policy rate and forms the actor observation in Sec.~\ref{sec:walk_policy}: angular velocity, projected gravity, joint state received from the motors, and the user command. A TensorFlow Lite graph of the trained actor runs on the phone CPU or GPU. The app sends joint-position targets over Wi-Fi (UDP) to a small onboard computer that forwards them to the XL430 bus. The same path supports walk, standup, and heading.

\paragraph{Communication and language.}
Wi-Fi carries motor state and commands. It also connects the phone to a laptop AI backend: the phone records speech on the microphone, the backend runs local speech-to-text (whisper.cpp) and a large language model, and the reply is spoken on the phone speaker with Piper text-to-speech. The API key stays on the laptop, not in the Android package.

\paragraph{Vision and human interaction.}
The rear camera runs on-device object detection and human pose estimation (ML Kit and MediaPipe). These detections support person following and other look-at behaviors together with the heading policy. The same camera and IMU also feed ARCore visual--inertial SLAM for onboard localization. The touchscreen displays status, camera overlays, and controls, so a user can operate the robot without a separate operator computer.

\section{Experiments}
\label{sec:experiments}

This section reports two groups of tests: locomotion training from scratch, and smartphone-application behavior on four phones.

\subsection{Locomotion}
\label{sec:exp_locomotion}

We compare the torque-aware walk policy trained with sagittal mirroring to the same policy trained without mirroring. Both methods use the environment, XL430 actuation model, observation split, action space, and reward weights of Sec.~\ref{sec:walk_policy}. The only intended difference is whether each PPO update concatenates a mirrored actor sample (Sec.~\ref{sec:mirroring}).

\begin{figure*}[htbp!]
\centering
\includegraphics[width=\textwidth]{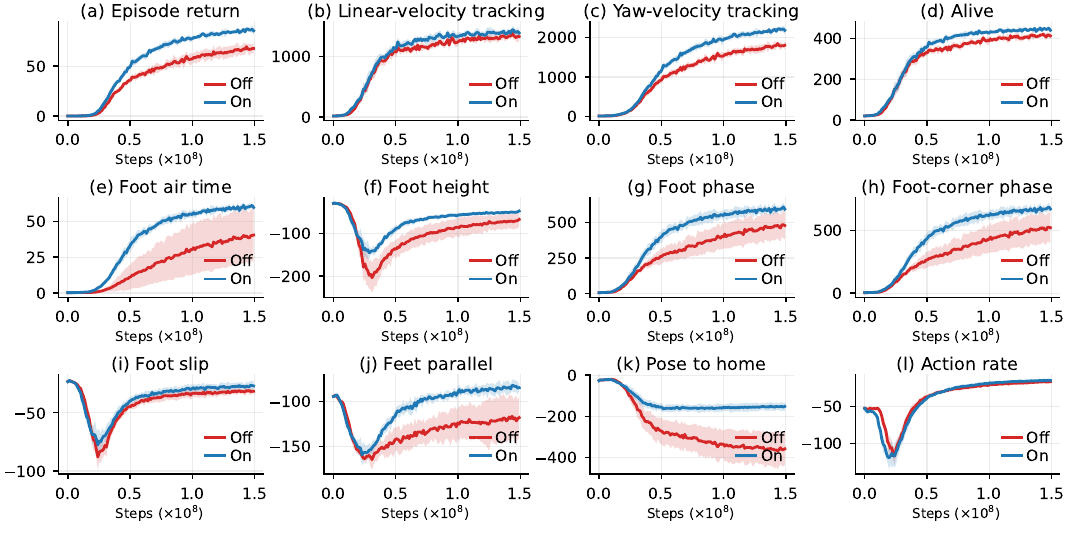}
\caption{Evaluation curves for the torque-aware walk policy trained from scratch with sagittal mirroring (blue) and without it (red). Both methods use the same reward weights from Table~\ref{tab:walk_rewards}. Each solid line is the mean over five seeds ($0$--$4$). The shaded region is $\pm 1$ standard deviation across seeds. Panel~(a) is the episode return: the weighted sum of the other terms, multiplied by the control timestep $\Delta t=0.02~\mathrm{s}$ and accumulated over the episode. Panels~(b)--(l) show the corresponding weighted terms before that $\Delta t$ factor. The horizontal axis is environment steps, shown up to $1.5\times 10^{8}$.}
\label{fig:locomotion_training}
\setcounter{subfigure}{0}
\refstepcounter{subfigure}\label{fig:loc_train_episode}
\refstepcounter{subfigure}\label{fig:loc_train_linvel}
\refstepcounter{subfigure}\label{fig:loc_train_angvel}
\refstepcounter{subfigure}\label{fig:loc_train_alive}
\refstepcounter{subfigure}\label{fig:loc_train_air}
\refstepcounter{subfigure}\label{fig:loc_train_height}
\refstepcounter{subfigure}\label{fig:loc_train_phase}
\refstepcounter{subfigure}\label{fig:loc_train_corner}
\refstepcounter{subfigure}\label{fig:loc_train_slip}
\refstepcounter{subfigure}\label{fig:loc_train_parallel}
\refstepcounter{subfigure}\label{fig:loc_train_pose}
\refstepcounter{subfigure}\label{fig:loc_train_action}
\end{figure*}

Each method is trained from scratch on flat terrain for five independent seeds $\{0,1,2,3,4\}$. The plots in Fig.~\ref{fig:locomotion_training} show the first $1.5\times 10^{8}$ environment steps. At every evaluation point the current policy is rolled out in $256$ evaluation environments. The solid line is the mean across the seeds that exist for that method. The shaded band is $\pm 1$ standard deviation across those seeds. The band is not the standard deviation inside a single evaluation.

The logged episode return is the accumulated per-step reward of Sec.~\ref{sec:walk_policy}: the weighted sum of the terms in Table~\ref{tab:walk_rewards}, multiplied by the control timestep $\Delta t=0.02~\mathrm{s}$ and then summed over the episode. The component panels show those same weighted terms before the $\Delta t$ factor. A larger value is better for a positive term. A value closer to zero is better for a cost term.

Fig.~\ref{fig:loc_train_episode} is the overall episode return. Mirroring raises this return earlier and to a higher final level. Its band is also narrower. The policy therefore finds a high-return gait with less dependence on the random seed. Training without mirroring is more seed-dependent and can remain in different local gait optima.

The two tracking terms explain much of that gap. Linear-velocity tracking (Fig.~\ref{fig:loc_train_linvel}) and yaw-velocity tracking (Fig.~\ref{fig:loc_train_angvel}) both increase earlier with mirroring, so the robot learns to follow the joystick command in fewer environment steps. The alive term (Fig.~\ref{fig:loc_train_alive}) increases as well, which means the mirrored policy spends more of the episode upright and terminates less often by falling.

The swing terms show how that command tracking is realized. Foot air time (Fig.~\ref{fig:loc_train_air}) and foot-height cost (Fig.~\ref{fig:loc_train_height}) rise (or recover toward zero) quickly and with a narrow band when mirroring is enabled. Without mirroring the same terms have a wide band: some seeds learn to lift the swing foot, while others keep the foot near the floor and drag it. That spread is a direct sign of seed-sensitive local optima.

Once the foot leaves the ground, mirroring also follows the scheduled gait more closely. Foot-phase tracking (Fig.~\ref{fig:loc_train_phase}) and foot-corner phase tracking (Fig.~\ref{fig:loc_train_corner}) are higher and more consistent with mirroring. Support is more stable as well: the slip cost (Fig.~\ref{fig:loc_train_slip}) returns closer to zero, so the stance foot slides less on the ground. The feet-parallel cost (Fig.~\ref{fig:loc_train_parallel}) also recovers closer to zero, which means the soles stay flatter to the floor. The pose cost (Fig.~\ref{fig:loc_train_pose}) remains nearer zero, which means the joints stay closer to the standing home pose instead of collapsing into a crouched or twisted stance.

Action rate is the one term that is not uniformly better at every step (Fig.~\ref{fig:loc_train_action}). Early in training the mirrored policy changes its action more than the baseline, so its action-rate cost is briefly larger. After that transient it reduces the action rate and finishes below the non-mirrored curve. The extra early motion is therefore paid back by a quieter late gait, not by a permanently more aggressive controller.

\begin{figure}[t]
\centering
\includegraphics[width=\columnwidth]{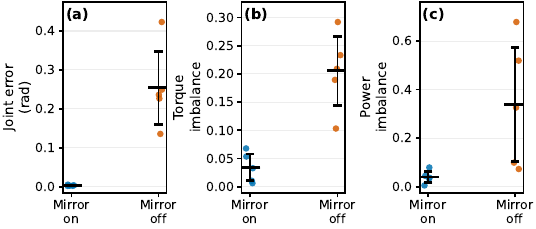}
\caption{Deterministic gait symmetry at approximately $4.5\times10^{8}$ environment steps. (a) Time-averaged joint-angle RMSE between one rollout and the sagittally mirrored phase-swapped rollout. (b) Relative left--right difference in mean absolute joint torque. (c) Relative left--right difference in mean absolute mechanical power $|\tau\dot{q}|$. Points are seeds; black markers show mean $\pm1$ standard deviation.}
\label{fig:locomotion_symmetry}
\end{figure}

We further evaluate the checkpoints after $5\times10^{8}$ environment steps in a nominal deterministic simulation, with domain randomization, noise, delay, backlash, joint offsets, and pushes disabled. Each policy receives a $0.5~\mathrm{m/s}$ forward command and is rolled out for $20~\mathrm{s}$ from the phase pairs $[0,\pi]$ and $[\pi,0]$ at $1.5~\mathrm{Hz}$; the first second is discarded. Fig.~\ref{fig:locomotion_symmetry} shows one point per training seed and the across-seed mean and standard deviation. Mirroring reduces the joint-angle symmetry error from $0.254$ to $0.003~\mathrm{rad}$, the left--right torque imbalance from $0.205$ to $0.034$, and the mechanical-power imbalance from $0.339$ to $0.040$.

\begin{figure} [htbp!]
    \centering
    \includegraphics[width=\linewidth]{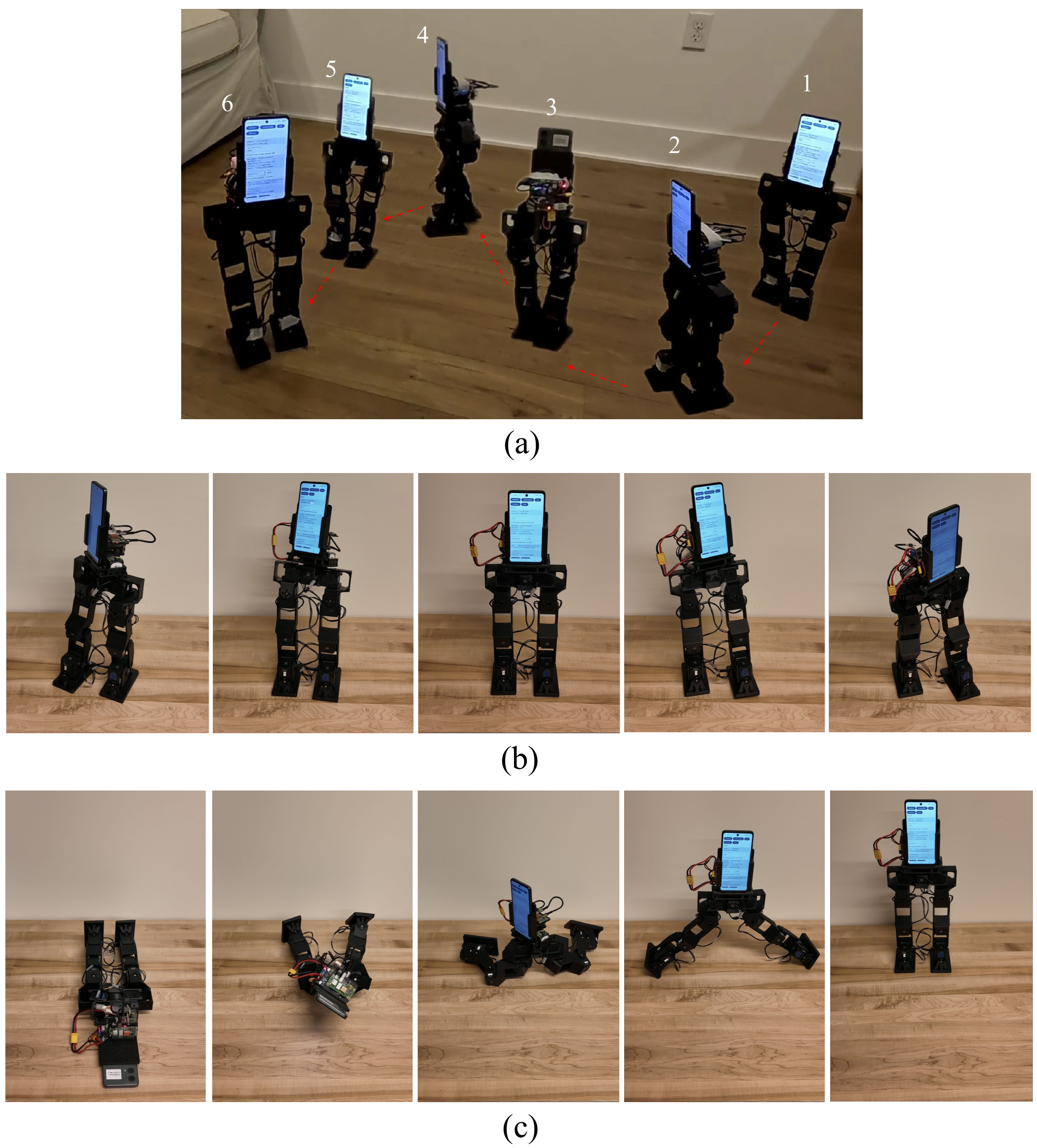}
    \caption{Physical demonstration of PhoneBot capabilities. (a)Bipedal locomotion. (b) Human tracking with front camera. (c) Stand up policy from face-down to fully upright stance.}
    \label{fig:timelines}
\end{figure}

The policies are executed on the physical robot. Fig.~\ref{fig:timelines} shows hardware tests of walking, human tracking and standing up.

\subsection{Smartphone Application}
\label{sec:exp_application}

The PhoneBot application is tested on four Android phones. All complete the same onboard functions: IMU sensing, UDP motor communication, locomotion-policy inference, on-device vision, and human--robot interaction. The set includes a 2017 Honor~9 and three prepaid phones bought for about \$50--\$80 and locked to Tracfone or Total Wireless (\href{https://www.amazon.com/dp/B0FTG1LVZR}{moto~g 2025}, \href{https://www.amazon.com/dp/B0FY2RTZ8Q}{Galaxy A16 5G}, and \href{https://www.amazon.com/dp/B0F7M5FSRT}{moto~g 2024}). The Honor~9 is an old flagship; the prepaid units are current low-cost phones. Both classes run the app without a separate vision or SLAM computer. That is the point of using the phone as the robot computer: even a cheap or old handset has enough CPU, camera, and IMU for these tasks. The current stack is CPU-bound, so GPU details are omitted.

Table~\ref{tab:phone_combined_rows} lists one phone per row. Hardware fields occupy the left block. The right block is CPU usage. Idle is the app with no function running. The other columns are IMU sensing, policy inference, vision detection, SLAM, and human--robot interaction. ARCore SLAM on the Honor~9 is marked $\times$ because Google services are unavailable on that Huawei device in US. In qualitative walking trials, ARCore SLAM remained stable despite the periodic body motion, with no evident degradation caused by the bipedal gait.

\begin{table*}[htbp!]
\centering
\caption{Hardware and onboard CPU usage of the four test phones. GPU is omitted because the current application stack is CPU-bound. Idle is the app with no function running. SLAM on Honor~9 is marked $\times$ because Google services are unavailable on that Huawei device. Prepaid prices are carrier-locked listings (Sep.\ 2026) except Honor~9, which is a used 2017 unit (China launch about \$340).}
\label{tab:phone_combined_rows}
\renewcommand{\arraystretch}{1.25}
\setlength{\tabcolsep}{3.5pt}
\small
\begin{tabular*}{\textwidth}{@{\extracolsep{\fill}}lccccccccccc@{}}
\toprule
 & \multicolumn{5}{c}{Hardware} & \multicolumn{6}{c}{CPU usage} \\
\cmidrule(lr){2-6} \cmidrule(lr){7-12}
Phone & Price & Released & CPU & RAM & Camera & Idle & \shortstack{IMU\\$200$~Hz} & Policy & \shortstack{Vision\\detection} & SLAM & HRI \\
\midrule
Honor 9 & used (2017) & Jul.\ 2017 & Kirin 960 & 4/6~GB & $12{+}20$~MP & $1.1\%$ & $12.1\%$ & $9.8\%$ & $17.9\%$ & $\times$ & $2.5\%$ \\
moto~g 2025 & \$50 & Jan.\ 2025 & Dimensity 6300 & 4~GB & $50{+}2$~MP & $0.4\%$ & $13.4\%$ & $10.3\%$ & $15.6\%$ & $34.3\%$ & $3.6\%$ \\
Galaxy A16 5G & \$50--\$80 & Oct.\ 2024 & Exynos 1330 & 4~GB & $50{+}5{+}2$~MP & $0.8\%$ & $8.5\%$ & $7.1\%$ & $17.6\%$ & $28.3\%$ & $5.2\%$ \\
moto~g 2024 & \$50--\$90 & Mar.\ 2024 & \shortstack{Snapdragon\\4 Gen 1} & 4~GB & $50{+}2$~MP & $0.8\%$ & $9.1\%$ & $6.3\%$ & $16.7\%$ & $25.2\%$ & $4.3\%$ \\
\bottomrule
\end{tabular*}
\end{table*}

\section{Conclusion}

This paper presented PhoneBot, a low-cost, open-source compact humanoid platform that integrates a commodity smartphone with a modular 13-DoF body for sensing, onboard computation, and interaction. Its locomotion pipeline combines a torque-aware actuator model with reinforcement learning in MJX. Evaluations across five random seeds showed that sagittal mirroring improved sample efficiency and training consistency compared with training without mirroring. Hardware demonstrations included walking, standing up, and camera-heading control, while application tests across four Android phones demonstrated the feasibility of reusing consumer smartphones for robot sensing, policy inference, and interaction. The hardware and software are released to support reproducible experimentation in education and research.

The current platform lacks arms, and its locomotion policies were trained on flat terrain. The limited torque of its low-cost servos also constrains payload and dynamic motion. Future work will investigate locomotion on uneven terrain and the integration of lightweight arms for object manipulation. Extending the training pipeline to these tasks will require accounting for changes in whole-body dynamics, contact conditions, and actuator demands. We also plan to explore language-model-based task planning that translates spoken instructions and visual observations into commands for learned skills. Deploying suitable language models on the smartphone could further reduce reliance on the external backend currently used for dialogue.
{
\bibliographystyle{IEEEtran}
\bibliography{references}

@article{toddlerbot,
  title={Toddlerbot: Open-source ml-compatible humanoid platform for loco-manipulation},
  author={Shi, Haochen and Wang, Weizhuo and Song, Shuran and Liu, C Karen},
  journal={arXiv preprint arXiv:2502.00893},
  year={2025}
}

@inproceedings{openbot,
  title={Openbot: Turning smartphones into robots},
  author={M{\"u}ller, Matthias and Koltun, Vladlen},
  booktitle={2021 IEEE International Conference on Robotics and Automation (ICRA)},
  pages={9305--9311},
  year={2021},
  organization={IEEE}
}

@misc{berkeley_humanoid_lite,
      title={Demonstrating Berkeley Humanoid Lite: An Open-source, Accessible, and Customizable 3D-printed Humanoid Robot}, 
      author={Yufeng Chi and Qiayuan Liao and Junfeng Long and Xiaoyu Huang and Sophia Shao and Borivoje Nikolic and Zhongyu Li and Koushil Sreenath},
      year={2025},
      eprint={2504.17249},
      archivePrefix={arXiv},
      primaryClass={cs.RO},
      url={https://arxiv.org/abs/2504.17249}, 
}

@inproceedings{liu2022bruce,
  author    = {Yeting Liu and Junjie Shen and Jingwen Zhang
               and Xiaoguang Zhang and Taoyuanmin Zhu
               and Dennis Hong},
  title     = {Design and Control of a Miniature Bipedal Robot
               with Proprioceptive Actuation for Dynamic Behaviors},
  booktitle = {2022 International Conference on Robotics
               and Automation (ICRA)},
  year      = {2022},
  pages     = {8547--8553},
  doi       = {10.1109/ICRA46639.2022.9811790}
}

@misc{mini_pi,
  author       = {{High Torque Robotics}},
  title        = {{Mini Pi High-Performance Bipedal Robot Platform}},
  howpublished = {\url{https://www.hightorquerobotics.com/mini-pi}},
  year         = {2024}
}

@misc{robotis_op3,
  author       = {{ROBOTIS}},
  title        = {{ROBOTIS OP3 Open Source Humanoid Robot}},
  howpublished = {\url{https://robotis.us/products/robotis-op3us}},
  year         = {2024},
  note         = {Accessed: Feb. 24, 2026}
}

@article{li2025reinforcement,
  author  = {Li, Zhongyu and Peng, Xue Bin and Abbeel, Pieter and Levine, Sergey and Berseth, Glen and Sreenath, Koushil},
  title   = {Reinforcement Learning for Versatile, Dynamic, and Robust Bipedal Locomotion Control},
  journal = {The International Journal of Robotics Research},
  volume  = {44},
  number  = {5},
  pages   = {840--888},
  year    = {2025}
}

@article{radosavovic2024real,
  author  = {Radosavovic, Ilija and Xiao, Tete and Zhang, Bike and Darrell, Trevor and Malik, Jitendra and Sreenath, Koushil},
  title   = {Real-World Humanoid Locomotion with Reinforcement Learning},
  journal = {Science Robotics},
  volume  = {9},
  number  = {89},
  pages   = {eadi9579},
  year    = {2024},
  doi     = {10.1126/scirobotics.adi9579}
}

@inproceedings{liao2025berkeley,
  author    = {Liao, Qiayuan and Zhang, Bike and Huang, Xuanyu and Huang, Xiaoyu and Li, Zhongyu and Sreenath, Koushil},
  title     = {Berkeley Humanoid: A Research Platform for Learning-Based Control},
  booktitle = {IEEE International Conference on Robotics and Automation (ICRA)},
  year      = {2025}
}

@misc{rudin2022learning,
      title={Learning to Walk in Minutes Using Massively Parallel Deep Reinforcement Learning}, 
      author={Nikita Rudin and David Hoeller and Philipp Reist and Marco Hutter},
      year={2022},
      eprint={2109.11978},
      archivePrefix={arXiv},
      primaryClass={cs.RO},
      url={https://arxiv.org/abs/2109.11978}, 
}

@inproceedings{zakka2025mujoco,
  author    = {Zakka, Kevin and Tabanpour, Baruch and Liao, Qiayuan and Haiderbhai, Mustafa and Holt, Samuel and Luo, Jing Yuan and Allshire, Arthur and Frey, Erik and Sreenath, Koushil and Kahrs, Lueder A. and Sferrazza, Carmelo and Tassa, Yuval and Abbeel, Pieter},
  title     = {Demonstrating {MuJoCo Playground}},
  booktitle = {Robotics: Science and Systems (RSS)},
  year      = {2025},
  doi       = {10.15607/RSS.2025.XXI.020}
}

@article{mittal2023orbit,
  author  = {Mittal, Mayank and Yu, Calvin and Yu, Qinxi and Liu, Jingzhou and Rudin, Nikita and Hoeller, David and Yuan, Jia Lin and Singh, Ritvik and Guo, Yunrong and Mazhar, Hammad and Mandlekar, Ajay and Babich, Buck and State, Gavriel and Hutter, Marco and Garg, Animesh},
  title   = {{Orbit}: A Unified Simulation Framework for Interactive Robot Learning Environments},
  journal = {IEEE Robotics and Automation Letters},
  volume  = {8},
  number  = {6},
  pages   = {3740--3747},
  year    = {2023}
}

@article{gu2024humanoid,
  author  = {Gu, Xinyang and Wang, Yen-Jen and Chen, Jianyu},
  title   = {{Humanoid-Gym}: Reinforcement Learning for Humanoid Robot with Zero-Shot {Sim2Real} Transfer},
  journal = {arXiv preprint arXiv:2404.05695},
  year    = {2024}
}

@inproceedings{tanaka2025mechanical,
  author    = {Tanaka, Yusuke and Zhu, Alvin and Wang, Quanyou and Hong, Dennis},
  title     = {Mechanical Intelligence-Aware Curriculum Reinforcement Learning for Humanoids with Parallel Actuation},
  booktitle = {2025 IEEE-RAS 24th International Conference on Humanoid Robots (Humanoids)},
  year      = {2025},
  pages     = {882--889},
  doi       = {10.1109/Humanoids65713.2025.11203130}
}

@inproceedings{nahrendra2023dreamwaq,
  author    = {Nahrendra, I Made Aswin and Yu, Byeongho and Myung, Hyun},
  title     = {{DreamWaQ}: Learning Robust Quadrupedal Locomotion with Implicit Terrain Imagination via Deep Reinforcement Learning},
  booktitle = {IEEE International Conference on Robotics and Automation (ICRA)},
  pages     = {5078--5084},
  year      = {2023}
}

@inproceedings{mittal2024symmetry,
  author    = {Mittal, Mayank and Rudin, Nikita and Klemm, Victor and Allshire, Arthur and Hutter, Marco},
  title     = {Symmetry Considerations for Learning Task Symmetric Robot Policies},
  booktitle = {IEEE International Conference on Robotics and Automation (ICRA)},
  pages     = {7433--7439},
  year      = {2024}
}

@article{scianca2020mpc,
  author  = {Scianca, Nicola and De Simone, Daniele and
             Lanari, Leonardo and Oriolo, Giuseppe},
  title   = {{MPC} for Humanoid Gait Generation: Stability and Feasibility},
  journal = {IEEE Transactions on Robotics},
  year    = {2020},
  volume  = {36},
  number  = {4},
  pages   = {1171--1188},
  doi     = {10.1109/TRO.2019.2958483}
}

@inproceedings{liu2025kidcosmo,
  author    = {Liu, Havel and Zhu, Mingzhang and
               Flores Alvarez, Arturo Moises and Lo, Yuan Hung and
               Ku, Conrad and Parres, Federico and Quan, Justin and
               Togashi, Colin and Navghare, Aditya and Wang, Quanyou and
               Hong, Dennis W.},
  title     = {From Screen to Stage: {Kid Cosmo}, A Life-Like,
               Torque-Controlled Humanoid for Entertainment Robotics},
  booktitle = {2025 IEEE-RAS 24th International Conference on
               Humanoid Robots (Humanoids)},
  year      = {2025},
  pages     = {677--684},
  doi       = {10.1109/Humanoids65713.2025.11203026}
}

@inproceedings{su2024leveraging,
  author    = {Su, Zhi and Huang, Xiaoyu and Ordo{\~n}ez-Apraez, Daniel and Li, Yunfei and Li, Zhongyu and Liao, Qiayuan and Turrisi, Giulio and Pontil, Massimiliano and Semini, Claudio and Wu, Yi and Sreenath, Koushil},
  title     = {Leveraging Symmetry in {RL}-Based Legged Locomotion Control},
  booktitle = {IEEE/RSJ International Conference on Intelligent Robots and Systems (IROS)},
  year      = {2024}
}
}

\end{document}